\documentclass[letterpaper, 10 pt, conference]{ieeeconf}
\IEEEoverridecommandlockouts

\usepackage{cite}
\usepackage{amsmath,amssymb,amsfonts}
\usepackage{algorithmic}
\usepackage{graphicx}
\usepackage{textcomp}
\usepackage{xcolor}
\let\labelindent\relax
\usepackage{enumitem}

\usepackage[caption=false,font=footnotesize]{subfig}

\usepackage{acronym}
\acrodef{hmd}[HMD]{head-mounted display}
\acrodef{vr}[VR]{Virtual Reality}
\newacro{ssq}[SSQ]{Simulator Sickness Questionnaire}
\acrodef{ros}[ROS]{Robot Operating System}
\acrodef{nasa-tlx}[NASA-TLX]{Nasa Task Load Index}
\acrodef{sc}[SC]{Shared-Control}
\acrodef{cs}[CS]{Control Switching}
\acrodef{sus}[SUS]{System Usability Scale}

\def\BibTeX{{\rm B\kern-.05em{\sc i\kern-.025em b}\kern-.08em
    T\kern-.1667em\lower.7ex\hbox{E}\kern-.125emX}}

\newcommand{\reals}{\mathbb{R}}
\begin{document}
\title{\LARGE \bf
Sharing but Not Caring:\\ Similar Outcomes for Shared Control and Switching Control in Telepresence-Robot Navigation\\
\thanks{© 2025 IEEE. Personal use of this material is permitted. Permission 
from IEEE must be obtained for all other uses, in any current or future 
media, including reprinting/republishing this material for advertising or 
promotional purposes, creating new collective works, for resale or 
redistribution to servers or lists, or reuse of any copyrighted 
component of this work in other works}

\thanks{This work was supported by the European Research Council (project ILLUSIVE 101020977) and Academy of Finland (projects BANG! 363637, CHiMP 342556).}
}

\author{Juho Kalliokoski$^{1}$, Evan G. Center$^{1}$, Steven M. LaValle$^{1}$, Timo Ojala$^{1}$, Basak Sakcak$^{1,2}$
\thanks{$^1$ Faculty of Information Technology and Electrical Engineering, University of Oulu, P.O. Box 4500, Oulu, FI-90014, Finland {\tt \footnotesize firstname.lastname@oulu.fi}}
\thanks{$^2$ Department of Advanced Computing Sciences, Maastricht University, Netherlands {\tt\footnotesize basak.sakcak@maastrichtuniversity.nl}}
}
\maketitle

\begin{abstract}
Telepresence robots enable users to interact with remote environments, but efficient and intuitive navigation remains a challenge. In this work, we developed and evaluated a \emph{shared control} method, in which the robot navigates autonomously while allowing users to affect the path generation to better suit their needs. We compared this with \emph{control switching}, 
where users toggle between direct and automated control.
We hypothesized that shared control would maintain efficiency comparable to control switching while potentially reducing user workload. The results of two consecutive user studies (each with final sample of $n=20$) showed that shared control does not degrade navigation efficiency, but did not show a significant reduction in task load compared to control switching. Further research is needed to explore the underlying factors that influence user preference and performance in these control systems.
\end{abstract}


\section{Introduction}
Telepresence robots represent a class of robotic systems in which a mobile robot is equipped with a camera streaming live video to a display that is watched by a remote user. 
These robots have already found a wide domain of applications ranging from business meetings and factory tours to personal events such as graduations and weddings. Furthermore, immersive-telepresence robots allow a panoramic camera stream to be watched by a remote user via a \ac{hmd}. These systems carry great potential in improving the user interaction with the remote environment, which is found to be lacking in the commercial systems~\cite{stoll2018wait}. One of the key advantages of telepresence robots lies in their ability to navigate the remote environment without being anchored to a particular location. However, effective control of the robot that accommodates the diverse preferences and needs of remote users is a challenge.


Controlling a robot in a remote location falls on a spectrum between full manual operation and complete autonomy (see \cite{moniruzzaman2022teleoperation} for a survey on this different levels of remote mobile robot control). On one end, manual control gives users direct authority over every movement, but it can be tiring and requires continuous input~\cite{rae2017robotic}. On the other end, full autonomoy allows the robot to navigate independently, reducing workload but potentially making decisions that do not match with user's preference or expectations. In between these extremes, different control strategies aim to assist users while preserving their sense of control, but each approach introduces trade-offs.

Among these methods that allow sharing the robot control between a user and an autonomous control module, one approach is to use waypoint navigation, where a user selects intermediate goals rather than steering the robot continuously~\cite{baker2020towards}. However, this method does not allow fine control over movement. Another approach is safeguard-based control, where the controller intervenes only to prevent collisions\cite{krotkov1996safeguarded,luo2020teleoperation,jiang2014shared,fong2001safeguarded}. While this reduces the task load, it does not assist with general navigation or decision-making. An alternative is control switching, where users alternate between manual and autonomous modes, either by choice or based on predefined conditions~\cite{petousakis2020human}. This approach, seen in commercial telepresence robots such as the Double 3~\cite{double3}, allows flexibility but requires users to manage transitions effectively. Finally, policy or trajectory blending combines user input with the control commands coming from an autonomous control module (see for example~\cite{dragan2013policy,6859352,2020,8914558,wang2014adaptive}). These methods offer smoother integration of user intent but are often limited in scope: some do not guarantee complete safety~\cite{6859352,2020}, and some focus on robots that are not mobile platforms~\cite{dragan2013policy}. Additionally, none of them took into consideration communication delay, which can significantly degrade the user experience in real-world scenarios.

In the context of telepresence, a shared control method should not only assist navigation but do so in a way that enhances user comfort rather than diminishing it, and takes into account the diverse preferences of users. The users should be able to adjust the robot’s trajectory effortlessly, enabling them to avoid close proximity to others or move toward points of interest not accounted for by autonomous navigation.
To account for these additional requirements, in our previous work~\cite{kalliokoski2022hidwa}, we developed a shared control method that allowed users to influence a robot's motion while maintaining collision-free navigation. We compared this approach to an alternative, where users switched between manual and autonomous control. While it allowed users to affect robot motion without manual steering, participants reported difficulty due to the robot overriding user input once manual influence ceased, leading to a feeling of ``fighting against the robot.''


Motivated by these findings, we developed a new shared control method which integrates the user input into planning to mitigate the feeling of ``fighting against the robot,'' and furthermore, allows the user to adjust the robot’s speed. Our approach modifies the costmap used for planning, in real time, based on user input, allowing for longer lasting trajectory adjustments. Unlike our previous work, our new approach continuously adapts the robot's trajectory to better align with user preferences, helping to reduce conflicts between the user and the system. Compared to blending or switching methods, our approach integrates user feedback directly into the planner, maintaining consistent autonomous control while respecting user-driven adjustments.

To evaluate this method, we conducted two user studies. In the first experiment, participants primarily focused on controlling the robot’s motion. However, we realized that without any additional cognitive demands, users were fully engaged in navigation, making it difficult to assess how well the system supports natural task execution in a realistic telepresence scenario. To address this, we conducted a second experiment where participants were also tasked with spotting animals in a forest environment during navigation. This additional task introduced a more representative use case for telepresence, where users must divide their attention between navigation and their primary objective. While our results show no significant differences in preference or the task load compared to control switching, they provide valuable insights into the challenges of designing shared control for immersive telepresence robots.

The remainder of the paper introduces the proposed method (\S\ref{sec:shared_control}), followed by the description of the user study (\S\ref{sec:experiment}), and concludes with a discussion on the results (\S\ref{sec:discussion}). 



\section{Proposed Shared Control Method}
\label{sec:shared_control}
We consider a scenario in which a robot must navigate to a goal while avoiding obstacles. During navigation, a remote user can provide inputs to adjust the robot's path based on visual feedback coming from a camera attached to the robot. In the following, we formally describe the motion planning problem and our proposed solution to integrate user input.
\subsection{Motion planning problem}
The robot kinematics is given in discrete time by the model
\begin{equation}
\label{eqn:unicycle}
\begin{alignedat}{2}
x(k+1)&=x(k) + v(k)\cos\theta(k)\Delta t \\
y(k+1)&=y(k) + v(k)\sin\theta(k)\Delta t \\
\theta(k+1)&=\theta(k) + \omega(k)\Delta t,
\end{alignedat}
\end{equation}
in which $(x,y)$ is the robot position and $\theta$ is the orientation with respect to a global reference frame, the control input $u=(v, \omega)$ corresponds to the linear and angular velocities with respect to the robot-fixed reference frame. The robot configuration is expressed as $q=(x,y,\theta)$ and it is constrained in the set $Q \subset \mathbb{R}^2 \times S^1$. The control input is subject to actuation constraints and it takes part in a compact set of admissible controls $U \subset \mathbb{R}^2$. 

Let $E \subset \reals^2$ be the planar environment in which the robot is moving. The environment contains obstacles, which are open sets and subsets of $E$ that prohibit the robot from occupying certain configurations due to collisions. These obstacles change dynamically, and their information is available to the robot locally. Let $E_{\text{obs}}(k) \subset E$ be the union of all the obstacles known to the robot at time $k\,\Delta t$. The part of the environment free from obstacles is $E_{free}(k)=E\setminus E_{obs}(k)$. Consecutively, the set of configurations that the robot can be in without collisions is denoted by $Q_{free}(k)$. 

The motion planning problem is defined as finding a sequence of control inputs $\tilde{u}^*=(u^*(1), u^*(2), \dots, u^*(N))$ for some $N$ such that the resulting sequence of configurations obtained via \eqref{eqn:unicycle} satisfies for all $k=1,\dots, N+1$ that $q(k) \in Q_{free}(k)$, $q(0)=q_I$, $q(1)=q_G$, in which $q_I$ and $q_G$ are initial and goal configurations. Furthermore, $\tilde{u}^*$ minimizes 
\begin{equation}
    J(u(1), \dots, u(N)) = \sum_{k=1}^N \ell(q(k),u(k),d(k)),
    \label{eq:full_horizon_cost}
\end{equation}
in which $d$ is the user input, and $\ell$ is an appropriate cost function.




\subsection{Planner}
We consider that the motion planning of the robot is achieved by two interacting modules: the \emph{planner} and the \emph{controller}. The \emph{planner} is responsible for computing an optimal path to the goal with respect to an appropriate objective, for example, a shortest path. Note that typically the path is computed over $E_{free}$ in which case it may not be 
feasible with respect to the robot
kinematics given in \eqref{eqn:unicycle}. 
Consecutively, the \emph{controller} determines the control inputs for tracking the resulting path. 

\begin{figure*}[!t]
\centering
\subfloat[
]{\includegraphics[width=1.5in, trim={0.75in 0 0.75in 0}, clip]{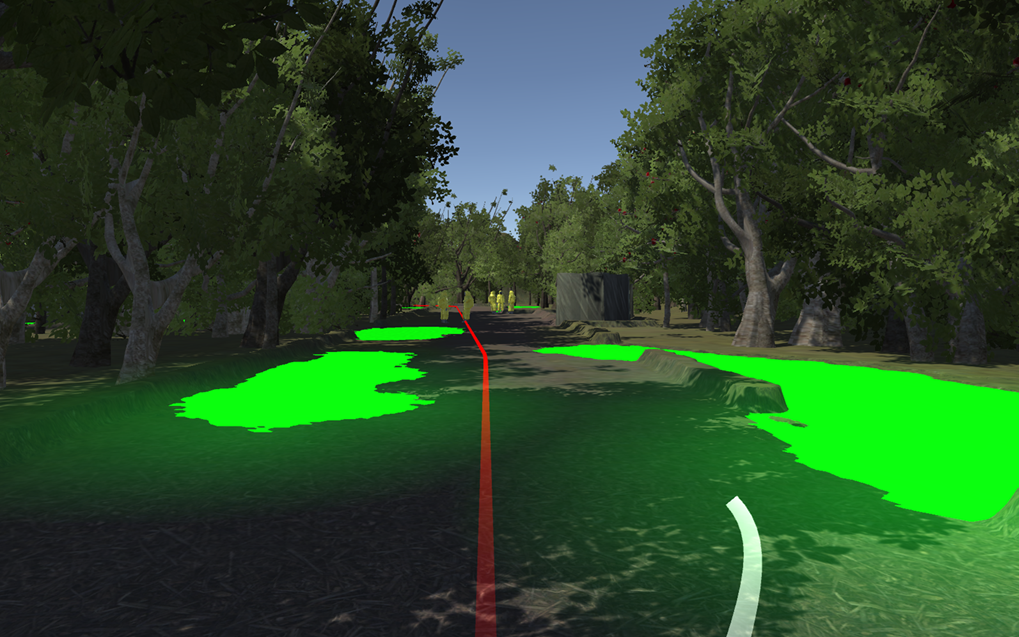}
\label{fig_user_input}}
\hfill
\subfloat[
]{\includegraphics[width=1.5in]{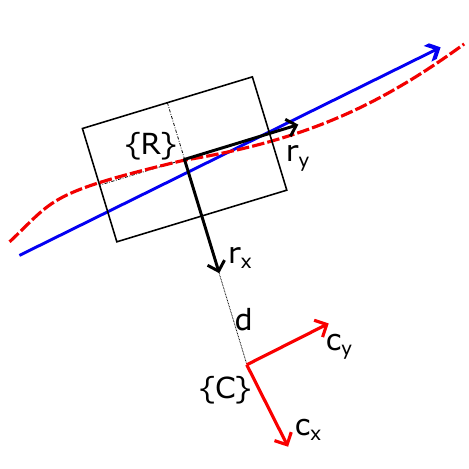}
\label{fig:c_frame}}
\hfill
\subfloat[
]{\includegraphics[width=1.5in]{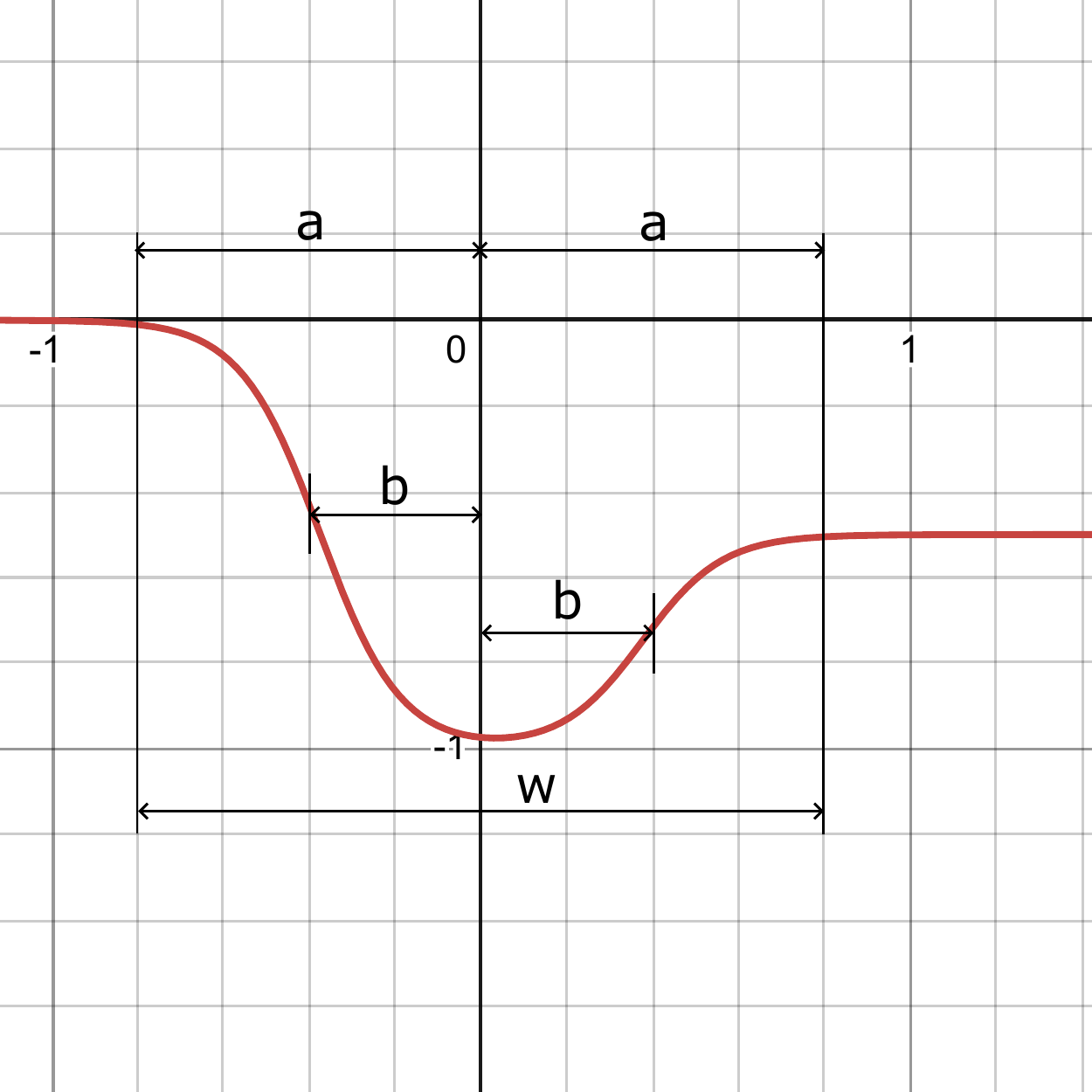}
\label{fig:sigmoid}}
\hfill
\subfloat[
]{\includegraphics[width=1.5in,trim={0.25in 0 0.75in 0.25in}, clip]{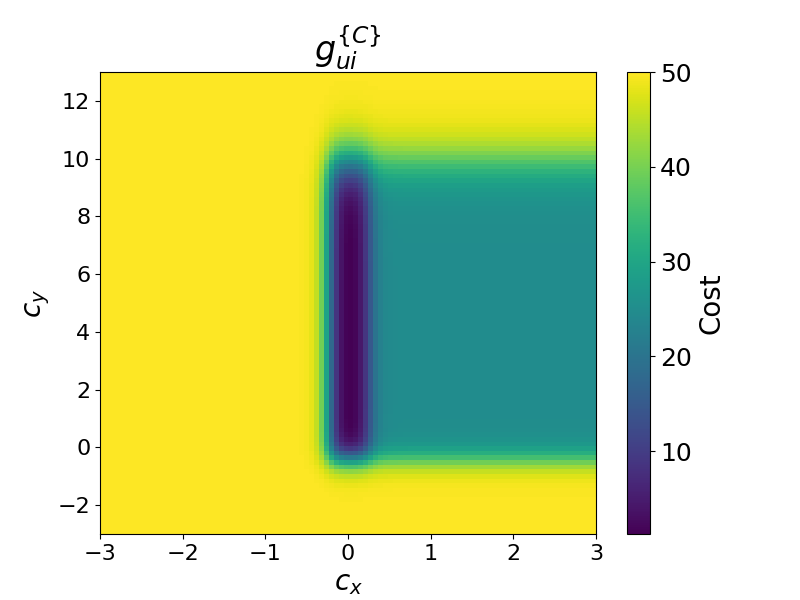}
\label{fig_costmap}}
\label{fig_ui_and_costmap}
\caption{(a) The user is presented with the robot path (red) and can select a lateral offset (white). (b) Robot and cost reference frames, $\{R\}$ and $\{C\}$, respectively. The origin of $\{C\}$ is $d$ away from the origin of $\{R\}$ and the $c_y$ axis is aligned with the vector along the line fitted (in blue) to the global path (red). (c) Plot of the lateral influence sigmoid functions with width $a\approx \frac{w}{2}$, and they are positioned symmetrically with offsets $\pm b$, in which $b=\frac{w}{4}$. (d) Heatmap showing a cost filter, $g_{ui}^{\{C\}}$, corresponding to the example given in (a).}
\vspace{-1.5em}
\end{figure*}

In our approach, we take the user input into account at the planner level. 
We assume that the user is 
presented with the path that the robot is following and when desired they can indicate a deviation from this path by adjusting the lateral offset through visual feedback (see Fig.\ref{fig_user_input}). 
Considering a planner based on grid search, we design a costmap that incorporates the desired deviation from the robot path. Therefore, the cost function given in \eqref{eq:full_horizon_cost} is implicitly minimized by the controller following an optimal path computed using the costmap.

Let $E_\Delta$ be a set of points representing the discretized position space. Given a user input $d$ and $E_{free}$, a cost function $g: E_\Delta \rightarrow \{0,1, \dots,255\}$, also referred to as a costmap, assigns a non-negative integer cost to each grid point. Finding an optimal path then amounts to applying an efficient grid search algorithm such as A*\cite{hart1968formal} on a 4 or 8 connected grid. In the next section, we explain the design of this costmap. To keep the notation light, we drop the dependency on time instance $k$, however it should be understood that the environment and the robot configuration is updated and the control input is determined accordingly.

\subsection{Costmap}
Our method integrates user input into the path planning by dynamically modifying the costmap used by the planner. 
It is composed of two components such that 
\begin{equation}
    g(x,y) = \max \Bigl(g_{obs}(x,y)\, ,\, g_{obs}(x,y) + g_{ui}(x,y) \Bigr),
    \label{eq:costmap_g}
\end{equation}
in which $g_{obs}$ is part of the cost function related to the obstacles and $g_{ui}$ is the one related to the user input. 

To take the robot size into account, the obstacles are inflated. Let $\bar{E}_{obs}$ be the inflated obstacles which typically corresponds to $\bar{E}_{obs}=E_{obs} \oplus \mathcal{D}$, in which $\mathcal{D}$ is an appropriate sized disk and $\oplus$ denotes the Minkowski sum. 
The grid points that fall into the obstacles, that is, $E_\Delta \cap \bar{E}_{obs}$, are assigned the highest value of 255 in the costmap, therefore 
\begin{equation}
     g_{obs}(x,y)= 
    \begin{cases}
    255 & \text{if } (x,y) \in \bar{E}_{obs}\\
    \texttt{decay}(x,y) & \text{otherwise},
    \end{cases}
    \label{eq:g_obs}
\end{equation}
in which \texttt{decay} is a function that assigns an integer cost to the points $E_\Delta \cap E_{free}$ that is inversely proportional to the distance from the nearest point in $\bar{E}_{obs}$ gradually decreasing from 255 to 0. 

To allow user influence over the robot's path, we introduce the cost component $g_{ui}$ such that minimizing this cost would lead to a robot motion according to the direction indicated by the user. To this end, the user input $d$ corresponds to a lateral offset along the $r_x$ axis of the robot-fixed reference frame (see Fig.~\ref{fig:c_frame}). Everytime a user input is received, a new reference frame $\{C\}$ is created with its origin at $(d, 0)$ (coordinates expressed with respect to $\{R\}$) and rotated so that the $c_y$ axis is aligned with the general direction of the robot. The general direction of the robot is determined as the vector 
aligned with the line fitted to a fixed-length portion\footnote{For this work we used a fixed-length of $5$ meters that is truncated as the robot is in the vicinity of the goal.} of the global path starting from the point closest to the robot's current position (see Fig.~\ref{fig:c_frame}).  

Once the reference frame for the cost, $\{C\}$, is defined, a cost filter that encodes the user input by creating a low-cost region which favors the direction indicated by the user is defined as follows 
\begin{equation}
    g^{\{C\}}_{ui}(c_x,c_y) = \texttt{round}\Bigl( s(1-f_{\text{lat}}(c_x)f_{\text{long}}(c_y)) \Bigr),
\end{equation}
in which $f_{\text{lat}}$ and $f_{\text{lon}}$ are functions that control the lateral and longitudinal influence of the user input over the cost filter, and $s \in [0,255]$ is a parameter adjusting the \emph{strength} of the cost filter that determines the contribution of the user input to the computation of the global path. Finally, to be consistent with the memory constraints on storing the cost filter, it is quantized into $256$ channels by rounding it to the nearest integer, that is, applying the \texttt{round} function. 

Specifically, the cost filter generates a \emph{cost valley} favoring the user indicated direction (see Fig.~\ref{fig_costmap}). The lateral influence, controlling the width of the cost valley is defined according to the following equation 
 \begin{equation}
                f_{\text{lat}}(c_x) = \frac{1}{1+e^{\frac{16}{w}(\frac{w}{4}+c_x)}} + \frac{p}{1+e^{\frac{16}{w}(\frac{w}{4}-c_x)}}-1,
\end{equation}
where 
$w$ is the width of the cost valley, and $p$ is a multiplier that lowers the cost on the side of the cost filter that aligns with the direction initially indicated by the user (see Fig.\ref{fig_costmap}). 
This adjustment encourages the robot to navigate on the side that matches the user's initial input when traversing the center of the valley is not possible.
Each sigmoid function has a width $a$, and the coefficient $16$ is chosen so that $a\approx \frac{w}{2}$. The coefficient $4$ appears because the sigmoids are offset symmetrically by a distance $b=\frac{w}{4}$ from the origin, ensuring that the valley centers around $c_x=0$ (see Fig.~\ref{fig:sigmoid}).
Consecutively, the longitudinal influence is determined by another set of sigmoid functions according to the following equation
\begin{equation}
    f_{\text{lon}}(c_y) = \frac{1}{\bigl(1+e^{2c_y-2l}\bigr)\, \bigl(1+e^{-4c_y-2}\bigr)}, 
\end{equation}
in which, $l$ corresponds the length of the cost valley, that is, the extent to which user input would affect the path along the direction of motion. 

Because the cost filter is defined in the reference frame $\{C\}$, we obtain the respective values in the global reference frame as 
\begin{equation*}
    g_{ui}(x,y)=g_{ui}^{\{C\}}\bigl(T_{\{C\}}(x,y)\bigr),
\end{equation*}
in which $T_{\{C\}}$ maps the points expressed in the global reference frame to those expressed in $\{C\}$. Finally, $g_{ui}$ is combined with $g_{obs}$, defined in \eqref{eq:g_obs}, according to $\eqref{eq:costmap_g}$ leading to the planner to return paths that are aligned with the user's input while ensuring obstacle avoidance.

\section{Experimental Evaluation}\label{sec:experiment}
We designed a user study to evaluate the effectiveness of our shared control system in an immersive telepresence setting. The study aimed to compare our approach with a conventional control switching method, assessing factors such as user workload, perceived control, navigation efficiency, and overall user experience.
By analyzing user behavior, questionnaire responses, and objective performance metrics, we aimed to understand whether our shared control method improves the telepresence experience and to identify potential challenges in its adoption.

\subsection{Design}

\begin{figure}[h]
\centering
\includegraphics[width=0.8\columnwidth,trim={0 3.5cm 0 1.5cm}, clip]{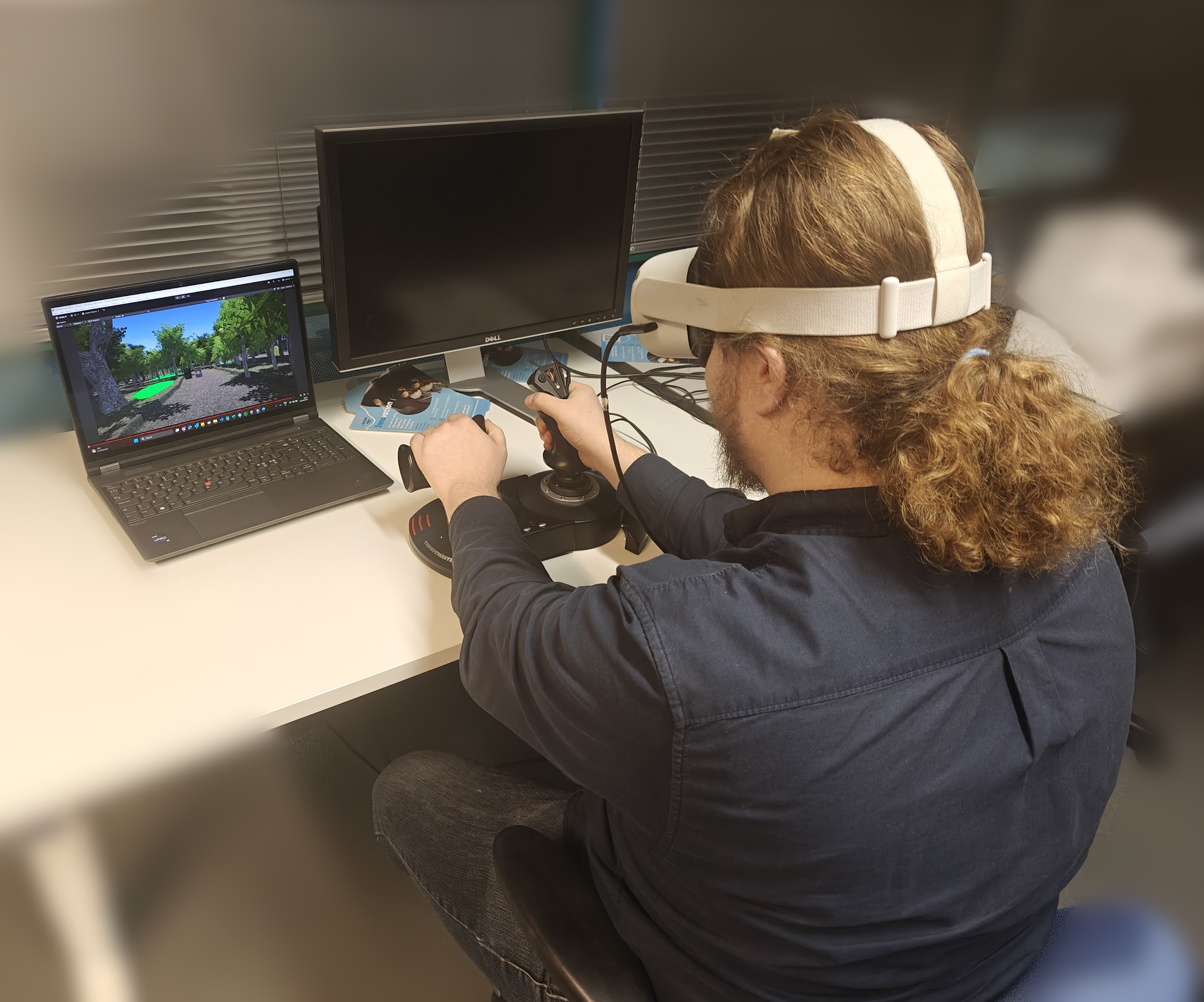}
\vspace{-0.75em}
\caption{A user controlling the simulated robot using a joystick.}
\label{fig:participant}
\vspace{-1.15em}
\end{figure}

In this study, we compared the proposed shared control method (\emph{shared control}) condition against a baseline controller (\emph{control switching}), in which the user is able to switch between completely manual, and completely autonomous control. 
\subsubsection{\textbf{Control Switching}} In the \ac{cs} condition, the robot moves autonomously, but the operator can take over the direct control of the robot by pressing a button allowing them to alter the course of the robot motion. The control input 
is determined as 
\begin{equation}
    u = 
    \begin{cases}
    u_h=(v_h,\omega_h) & \text{if user input given} \\
    u_a=(v_a,\omega_a) & \text{otherwise},
    \end{cases}
\end{equation}
where $u_a$ is the input given by the controller and $u_h$ is the user input given using the joystick that can be seen in Fig.~\ref{fig:participant}.
Let $(j_x,j_y) \in [-1,1]\times[-1,1]$ be the input given by the operator where $j_x$ is the horizontal position of the joystick with 0 being the origin, and $j_y$ is the position of the speed lever with 0 being the middle position. The user input is mapped to $u_h = (v_h,\omega_h)$ as follows:
\begin{equation}
    v_h=
    \begin{cases}
        1+0.5\,j_y & \text{if } j_y>0\\
        1+j_y & \text{otherwise},
    \end{cases}
    \label{eq:speed}
\end{equation}
and
\begin{equation}
    \omega_h=
    \begin{cases}
        j_x & \text{if } j_y>0\\
        j_x+0.8\,j_y& \text{otherwise}.
    \end{cases}
    \label{eq:omega}
\end{equation}
This mapping is designed to ensure intuitive control, where the robot moves at its intended velocity ($1 m/s$) when the speed lever is in the central position. Pushing the lever to its upper limit allows the robot to reach its maximum velocity ($1.5 m/s$) while pulling it to the lower limit brings it to a stop. 
The angular velocity mapping is designed to reduce the maximum angular velocity at lower speeds, minimizing the risk of accidental oversteering. However, the robot is still allowed to turn at a slow rate even when stationary.

To let the user adjust the speed while the robot was in autonomous mode, we limited the maximum linear velocity to the value given by \eqref{eq:speed}.

\subsubsection{\textbf{Shared Control}}
Similar to the CS 
condition, also in this case, the operator uses a joystick to provide input to the system. However, this time the robot would always follow the input $u_a=(v_a,\omega_a)$ given by the controller.

To modify the robot's path, the operators used the Shared Control method explained in Section~\ref{sec:shared_control}. To do this, they would hold down the trigger button and move the joystick left or right. The position of the joystick at the moment they released the trigger determined the placement of the costmap filter. The displacement $d$ of the costmap filter relative to the robot's center was computed as:
\begin{equation}
    d=5\,j_x,
\end{equation}
where $j_x$ is the horizontal axis value of the joystick input.
A negative $j_x$ value shifted the costmap filter to the left, while a positive value shifted it to the right relative to the robot's orientation.

To allow the users to adjust the robot's speed, the controller's input $u_a=(v_a,\omega_a)$ was limited such that:
\begin{equation}
    v_a \leq v_h, \quad |\omega_a| \leq \omega_h,
\end{equation}
where $v_h$ and $\omega_h$ are the user-defined maximum linear and angular velocities, respectively, as defined in \eqref{eq:speed} and \eqref{eq:omega}.

\subsection{Hypotheses}
To investigate the impact of different control methods on user experience and performance in immersive telepresence, we tested the following hypotheses:
\begin{enumerate}[label=\textbf{H\arabic*:}]
    \item The task load in the \emph{shared control} condition will be lower than in the \emph{control switching} condition as indicated by lower NASA-TLX scores.
    \item There will be no difference in user performance across conditions as measured by similar cumulative radiation values (see Section~\ref{sec:study_setup}), and similar completion times.
\end{enumerate}

\subsection{Experiment 1}

\subsubsection{Study setup} 
\label{sec:study_setup}
\begin{figure}[h]
\centering
\includegraphics[width=2.5in]{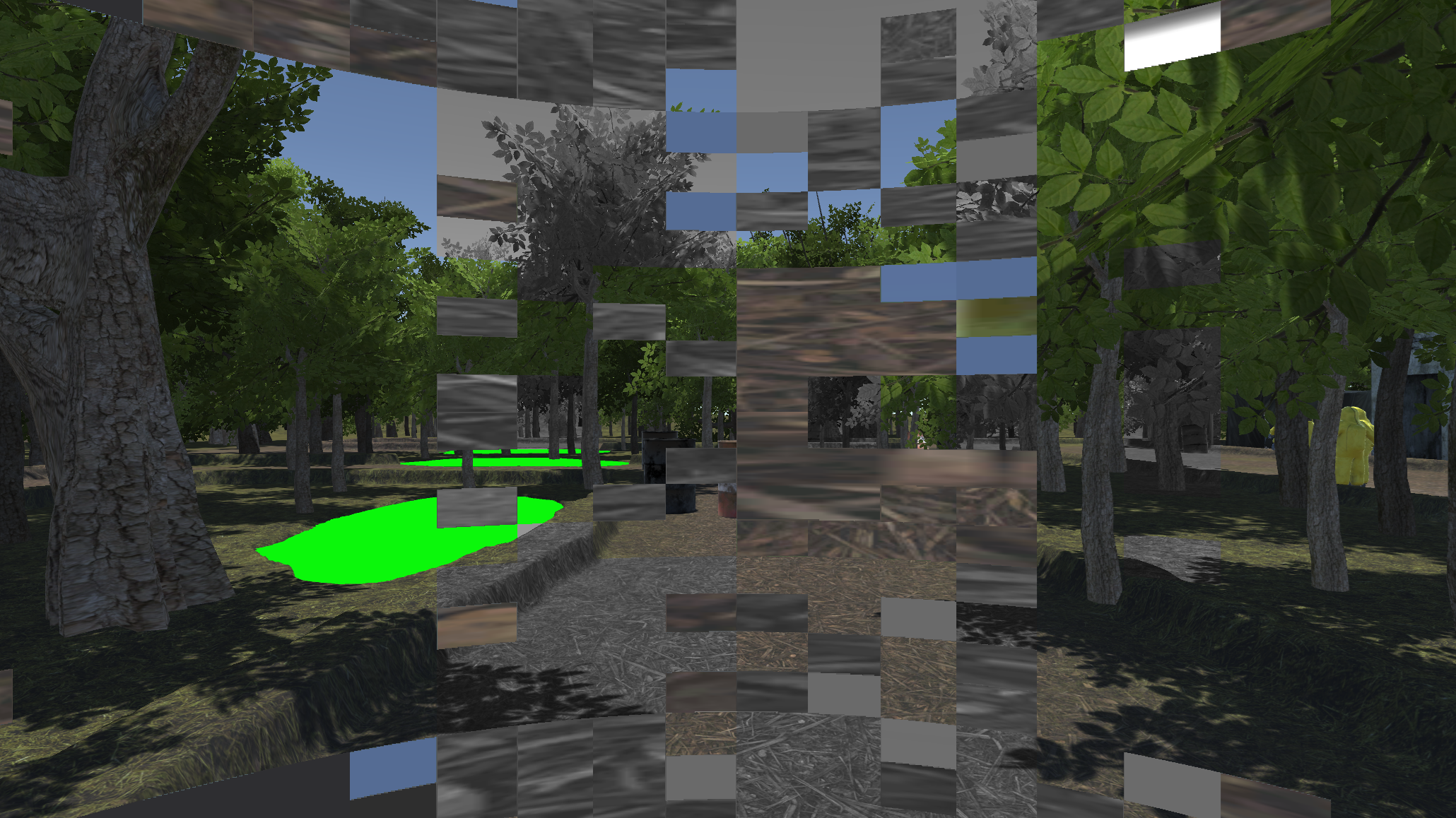}
\caption{Simulated video feed disruptions.}
\label{fig:disturpt}
\end{figure}

The hypotheses were tested using a simulated environment created with the Unity 3D-game engine. The virtual environment consisted of a forest area with path going through it (see Fig.~\ref{fig_user_input}). Participants encountered pools of toxic waste that they had to avoid, open areas with concrete buildings, and people in hazard suits, creating a structured yet varied setting. The simulated telepresence robot used in the experiment comprised of a mobile base as described in Section~\ref{sec:shared_control} and a simulated 360$^\circ$ camera attached 1.5 meters above the robot from which the users could see the virtual world using a virtual reality headset.
The robot's autonomous navigation was based on the \ac{ros} and its Nav2 project \cite{macenski2020marathon2}. In this experiment, the robot was navigating towards a fixed goal, but had no prior knowledge of the positions of the toxic pools or people standing along the path. It was the participants’ responsibility to avoid the toxicity coming from the pools as effectively as possible. Toxicity levels were visually represented using a green glow effect on the floor, with the intensity of the glow indicating the level of toxicity at a given location. To better simulate real-world telepresence scenarios, a $1$-second delay was introduced to the user controls, reflecting network latency that can occur in remote operation. Additionally, a visual interference effect was applied using a shader, overlaying the user's view with a distorted version of the environment to simulate video feed disruptions due to lost data (see Fig.~\ref{fig:disturpt}).
\subsubsection{Analyses} 
To measure the task load experienced during each of the conditions, we asked the participants to fill out \ac{nasa-tlx} questionnaire after completion of each of the tasks. The questionnaire consists of six subscales, each representing a different aspect of workload. To calculate the total workload, each subscale score is multiplied by a weight value. These weights are determined individually for each participant by having them compare all pairs of subscales and rank their relative importance to the task they just completed.

To measure the performance of the participants, we measured the completion time of the task, and the cumulative radiation during the task. The cumulative radiation was calculated by taking the distance to the closest toxic pool at each time step and integrating it over time.
As the participants in our previous study had mentioned having control over the robot as the main reason for preferring the \ac{cs} condition, we asked the participants Likert-scale rating for how much they felt like they had control over the robot.
To learn more about the usability of the methods, we asked the participants to fill out a \ac{sus}-questionnaire\cite{brooke1996sus} after each of the conditions.
After the second task, we additionally asked forced-choice questions about the participants' preference between the two methods. We also asked open-ended questions about the reasons for their choices and whether there were any specific situations in which they would prefer one of the methods over another.

Continuous data was tested for normality using Shapiro-Wilk tests. T-tests were used to evaluate normally distributed data and Wilcoxon signed rank tests were used for non-normal or Likert scale data. Bayes factors were calculated to estimate support for null hypotheses. For normally distributed data, the `ttestBF' function from the R package `BayesFactor' package was used \cite{moreyBayesFactor}, and for non-normally distributed data, the Wilcoxon-signed-rank Bayes factor approach from van Doorn et al. 2020 was used \cite{van2020bayesian}. Both cases implemented a standard Cauchy prior. Bayes factors between 1 and 0.33, less than 0.33, and less than 0.2 are interpreted here as anecdotal, moderate, and strong support for the null hypothesis, respectively. 
\subsubsection{Procedure}
The participants were presented with two conditions corresponding to the \ac{sc}, and \ac{cs} control methods. The same path was used in both conditions, however, on one of the conditions the path was traversed in reverse direction to mitigate learning effects. The order and direction of these conditions was counterbalanced across participants.
Upon arrival, the participants were greeted by a researcher and signed a form to indicate their consent to participate. Participants were then asked if they were experiencing any nausea or headaches to ensure they were not feeling unwell before the experiment began. Afterward, they filled out a demographic questionnaire and received general instructions. Their interpupillary distance was measured, and they were shown how to put on the \ac{hmd}.

After the experimenter made sure the \ac{hmd} was fitted properly and that the participant was able to see the virtual environment clearly, the participants were asked to take off the \ac{hmd}, and they were given the instructions for the first task. The participants were told that they were inspecting a forest where toxic waste had been dumped. Although the robot could navigate autonomously, it could not detect the toxic pools or people standing along the path. Participants were instructed to avoid the toxic waste indicated by the green glow on the ground. The control method for the first task was explained, after which they were told to put the \ac{hmd} back on and practice the controls in a separate environment for up to 5 minutes. After practice, the participants were asked if they were ready for the task, and the appropriate task scenario was initiated. Upon completing the task, participants removed the \ac{hmd} and filled out a questionnaire regarding their experience with that specific control method. This procedure was then repeated with the other control method. Finally, the participants were given 20€ gift card to a local store chain as compensation.

\subsubsection{Participants}
All participants gave written informed consent to participate in the experiments. All experimental procedures were in accordance with the Declaration of Helsinki and approved by the University of Oulu ethical review board (ERB). A total of 28 participants were recruited from the University of Oulu campus and community, but eight of them had to be excluded as they were not able to finish both of the tasks without taking off the \ac{hmd}. Of the 20 included participants, five were women and 15 were men, and their ages ranged from $21$ to $45$ years with a mean of $29.05$. The responses of the participants to how often they use \ac{vr} systems were: $35\%$ just a couple of times and at least once, $25\%$ once or twice a year, $35\%$ once or twice a month, and $5\%$ once or twice a week.
\subsubsection{Results}
\begin{figure*}[!t]
\centering
\subfloat[]{\includegraphics[width=2in]{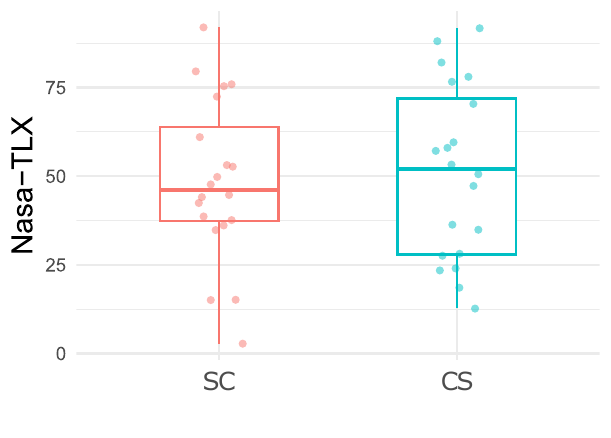}
\label{fig:pilot1_nasa}}
\hfil
\subfloat[]{\includegraphics[width=2in]{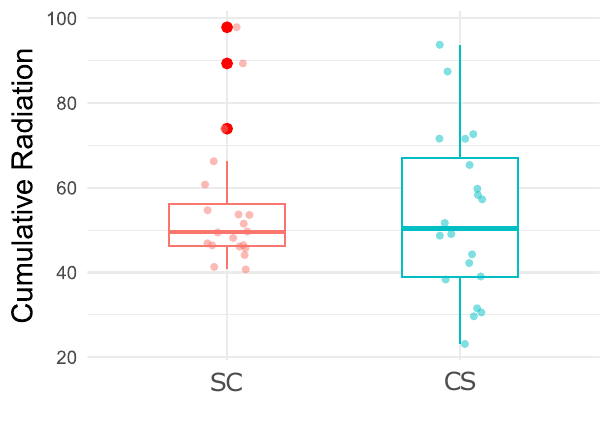}
\label{fig:pilot1_rad}}
\hfil
\subfloat[]{\includegraphics[width=2in]{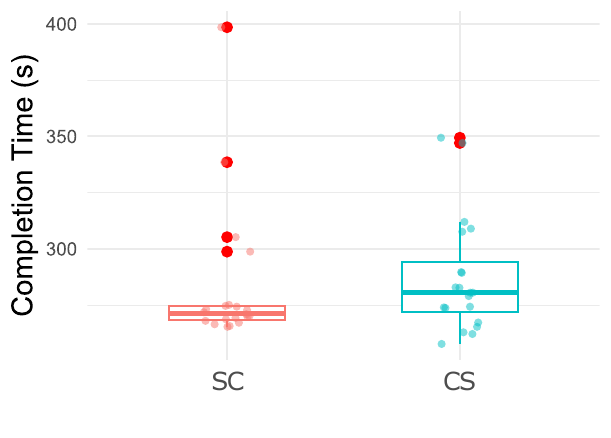}
\label{fig:pilot1_time}}
\caption{Results from Experiment 1: (a) Nasa-TLX scores, (b) Cumulative Radiation scores, (c) Completion times}
\label{fig:pilot1}
\end{figure*}
Distributions corresponding to NASA-TLX, cumulative radiation, and completion times in Experiment 1 can be seen in Figure~\ref{fig:pilot1}. The NASA-TLX score differences followed a normal distribution, as indicated by Shapiro-Wilk test, $W=0.97$, $p=0.66$. Therefore, a paired t-test was performed to compare the differences in task load scores between the \ac{sc} ($M=48.5$, $SD=23.04$) and \ac{cs} ($M=50.92$, $SD=24.58$) conditions.  The results indicated that there was no significant difference in the task load between the two conditions, $t(19)=-0.68$, $p=0.51$, $BF=0.21$. 

The differences in cumulative radiation scores also displayed normal distribution characteristics, as indicated by Shapiro-Wilk test, $W=0.95$, $p=0.29$. Therefore, a paired t-test was performed to compare the differences in cumulative radiation scores between the \ac{sc} ($M=55.35$, $SD=15.47$) and \ac{cs} ($M=53.31$, $SD=19.48$) conditions. Once again, the results indicated that there was no significant difference in the scores between the two conditions, $t(19)=-0.56$, $p=0.58$, $BF=0.20$. 

The the differences between the completion times were not normally distributed, as indicated by Shapiro-Wilk test, $W=0.90$, $p=0.039$. Since the normality assumption for the paired t-test was violated, a Wilcoxon signed-rank test was performed to compare the differences in the completion times between the \ac{sc} ($M=283.20$, $SD=32.40$) and \ac{cs} ($M=287.34$, $SD=25.80$) conditions. The results indicated that there was no significant difference in the scores between the two conditions, $Z=-0.78$, $p=0.45$, $BF=0.19$.

Both results show that the performance of the \ac{sc} is slightly better than \ac{cs} but there is no significant difference between the conditions, supporting H1.
When asked which condition they preferred more, 11 out of 20 participants ($55\%$) selected the \ac{cs} condition, showing no significant tendency in either direction in preference.

A Wilcoxon signed-rank test was performed to compare the differences in the feeling of control ratings between the \ac{sc} ($M=5.3$, $SD=0.98$) and \ac{cs} ($M=5.1$, $SD=1.17$) conditions. There was no significant difference in the feeling of control as indicated by a Wilcoxon signed-rank test, $Z=0.60$, $p=0.59$, $BF=0.20$.
We also asked ``How comfortable was altering the robot's path". Once again, a Wilcoxon signed-rank showed that there was no significant difference in  the comfort between the \ac{sc} ($M=4.6$, $SD=1.60$) and the \ac{cs} ($M=4.9$, $SD=1.21$), $Z=-0.79$, $p=0.47$, $BF=0.23$.

A Shapiro-Wilk test indicated that the \ac{sus} scores of the \ac{sc} ($M=58.63$, $SD=12.60$) and the \ac{cs} ($M=54.75$, $p=16.97$) methods are not normally distributed, $W=0.89$, $p=0.023$. There was no significant difference between the \ac{sus} scores as shown by A Wilcoxon signed-rank test, $Z=0.26$, $p=0.80$, $BF=0.19$.

\subsection{Experiment 2} 
One of the purposes of shared control is to reduce the effort required for navigation so that users can focus on other tasks. In Experiment 1, participants had no other task in addition to controlling the robot, making it difficult to assess this benefit. To better reflect the use of real-world telepresence, we conducted Experiment 2, where in addition to navigating according to the same instructions as in Experiment 1, the participants also had to spot animals while navigating. This allowed us to evaluate how well the shared control method supports multitasking.

\subsubsection{Study setup} 
Experiment 2 followed the same general setup as Experiment 1, with participants navigating a virtual forest environment using the telepresence robot. However, to better evaluate the benefits of shared control, an additional task was introduced: Participants were asked to report the animals they spotted during navigation. When they saw an animal in the environment, they had to scroll through pictures of animals using buttons on the controller and pull the trigger button when they found a matching picture from the list. This simulated a real-world scenario in which users must divide their attention between navigation and some other objective.

Another key difference from Experiment 1 was the way toxicity levels were conveyed. Instead of a visual green glow on the ground, participants were alerted to increasing toxicity levels through a Geiger counter sound effect, which intensified as they approached hazardous areas. This change was made to encourage participants to focus on both navigation and the animal-spotting task, rather than relying on direct visual cues for where the robot should be moving.

In Experiment 1, a visual interference effect was introduced to simulate video feed disruptions, illustrating a scenario where shared control could assist when visibility is compromised. However, to reduce unnecessary \ac{vr}-sickness and ensure that participants could fully concentrate on the animal-spotting task, this effect was omitted in the second experiment. The experiment procedure remained otherwise the same, with participants experiencing both control methods in a counterbalanced order, practicing the controls before the task, and filling out questionnaires after each condition.
\subsubsection{Analyses}
In addition to the measures used in Experiment 1, we asked participants to complete a locus-of-control questionnaire~\cite{craig1984scale}. This decision was based on the feedback from Experiment 1, where some participants emphasized the importance of feeling in control of the robot. We aimed to determine whether there was a connection between participants' preference for a control method and the personality traits measured by this questionnaire.

\subsubsection{Participants} 
A total of 24 participants were recruited from the University of Oulu campus and community. Participants from Experiment 1 were not allowed to participate in thise experiment.  Four participants had to be excluded as they were not able to complete both tasks without taking off the \ac{hmd}. Of the 20 included participants, $10$ were women, nine were men and one preferred not to say, and their ages ranged from $20$ to $46$ years with a mean of $27.75$.
The responses of the participants to how often they use \ac{vr} systems were: $30\%$ had never used \ac{vr}, $25\%$ just a couple of times and at least once, $20\%$ once or twice a year, $15\%$ once or twice a month, $5\%$ once or twice a week, and $5\%$ several times a week.
\subsubsection{Results}
\begin{figure*}[!t]
\centering
\subfloat[]{\includegraphics[width=2in]{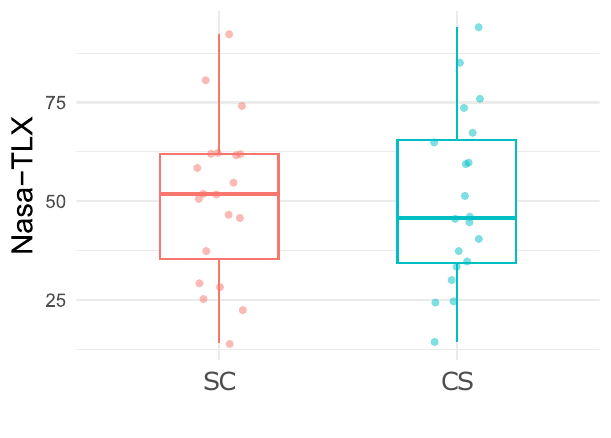}
\label{fig:pilot2_nasa}}
\hfil
\subfloat[]{\includegraphics[width=2in]{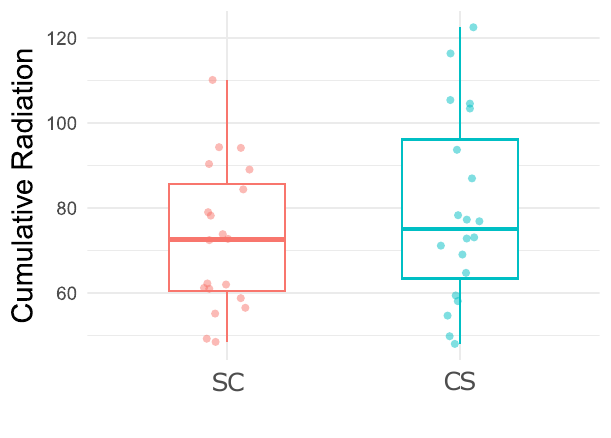}
\label{fig:pilot2_rad}}
\hfil
\subfloat[]{\includegraphics[width=2in]{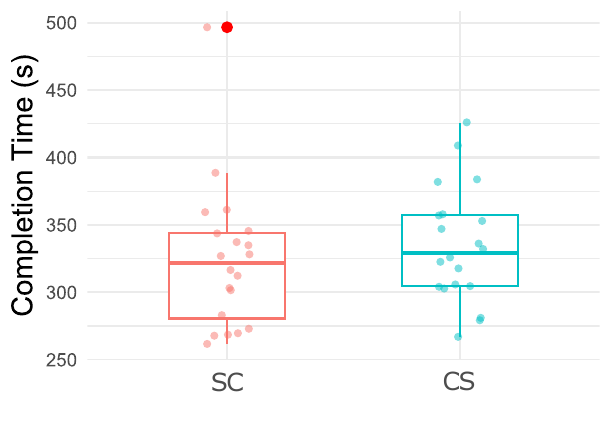}
\label{fig:pilot2_time}}
\caption{Results from Experiment 2: (a) Nasa-TLX scores, (b) Cumulative Radiation scores, (c) Completion times}
\label{fig:pilot2}
\vspace{-2em}
\end{figure*}

Distributions corresponding to NASA-TLX, cumulative radiation, and completion times in Experiment 2 can be seen in Figure~\ref{fig:pilot2}. This time the differences in NASA-TLX scores were not normally distributed, as indicated by Shapiro-Wilk test, $W=0.87$, $p=0.010$. Therefore, a Wilcoxon signed-rank test was performed to compare the differences in task load scores between the \ac{sc} ($M=50.58$, $SD=20.18$) and \ac{cs} ($M=50.33$, $SD=21.65$) conditions.  The results indicated that there was no significant difference in the task load between the two conditions, $Z=-0.75$, $p=0.47$, $BF=0.18$.

The Shapiro-Wilk test showed that the differences in the cumulative radiation scores were again normally distributed, $W=0.99$, $p=0.99$. Therefore, a paired t-test was performed to compare the differences in cumulative radiation scores between the \ac{sc} ($M=72.69$, $SD=17.02$) and \ac{cs} ($M=79.34$, $SD=21.98$) conditions. The results indicated that there was no significant difference in the scores between the two conditions, $t(19)=-1.23$, $p=0.23$, $BF=0.35$. 

The distribution of the difference between the completion times was normal, as indicated by Shapiro-Wilk test, $W=0.97$, $p=0.84$. A t-test was performed to compare the differences in the completion times between the \ac{sc} ($M=323.98$, $SD=54.40$) and \ac{cs} ($M=334.70$, $SD=42.72$) conditions. The results indicated that there was no significant difference in the scores between the two conditions, $t(19)=-0.85$, $p=0.41$, $BF=0.24$. 
When asked ``Thinking about both of the tasks you did, which control method did you prefer more?" 10 out of 20 participants ($50\%$) selected the \ac{cs} condition showing no significant tendency in either direction in preference.

We asked the participants for a Likert-scale rating on how much they felt like they had control over the robot. A Wilcoxon signed-rank test was performed to compare the differences in the feeling of control ratings between the \ac{sc} ($M=5.45$, $SD=1.15$) and \ac{cs} ($M=5.65$, $SD=0.75$) conditions. There was no significant difference in the feeling of control, $Z=-0.64$, $p=0.57$, $BF=0.20$.
We also asked ``How comfortable was altering the robot's path". A Wilcoxon signed-rank showed that there was no significant difference in comfort between the \ac{sc} ($M=5$, $SD=1.13$) and the \ac{cs} ($M=5.15$, $SD=1.35$), $Z=-0.19$, $p=0.87$, $BF=0.17$.

A Shapiro-Wilk test indicated that the \ac{sus} scores of the \ac{sc} ($M=62.25$, $SD=13.55$) and the \ac{cs} ($M=64.0$, $SD=10.11$) methods were normally distributed, $W=0.98$, $p=0.88$. There was no significant difference between the \ac{sus} scores as shown by a paired t-test, $t(19)=-0.84$, $p=0.41$, $BF=0.24$. 

The normality of the locus-of-control was tested using a Shapiro-Wilk test. The results showed that the data were normally distributed for both participants who preferred the \ac{sc} method ($M=29.4$, $SD=10.09$, $W=0.91$, $p=0.26$) and those who preferred the \ac{cs} method ($M=30.8$, $SD=9.41$, $W=0.95$, $p=0.67$).
A t-test was performed to compare Locus of Control scores between participants who preferred \ac{sc} and those who preferred \ac{cs} methods. The results indicated that there was no significant difference in Locus of Control scores between the two groups, $t(18)=-0.32$, $p=0.75$, $BF=0.32$. 

\section{Discussion}\label{sec:discussion}
Our study examined how a \emph{shared control} system compares to \emph{control switching} in immersive telepresence robot navigation. In \emph{Experiment 1}, participants navigated through a controlled environment using both methods to evaluate task performance and workload. Results showed that although the shared control did perform comparably to control switching, there were \emph{no significant differences} between the two approaches, suggesting that neither offered a clear advantage in terms of efficiency or user effort. In \emph{Experiment 2}, where participants had to divide their attention between navigation and an additional cognitive task, we observed a similar pattern: shared control maintained the performance levels comparable to switching, but did not significantly reduce workload. Furthermore, associated Bayes factors in nearly every case indicated moderate to strong support for null hypotheses, indicating a true lack of differences between conditions rather than merely a lack of statistical power to detect differences.

Given this outcome, the key question becomes why shared control did not demonstrate a stronger benefit over control switching. One possibility is that both methods allow users to manage navigation effectively, just through different means. Users who preferred control switching often cited a greater sense of control (\textit{``I feel more control on the robot"}) and reduced complexity, noting that the ability to manually toggle between control modes allowed them to feel more directly in charge of the robot’s movements (\textit{``It was easy to use. Less complex systems can target many users."}). Some also found that control switching mode helped reduce motion sickness (\textit{``Its easier to use and is more steady, gave me less motion sickness"}), possibly because it eliminated small, unintended adjustments that could contribute to visual instability.
On the other hand, those who favored shared control appreciated its ability to reduce effort and make navigation feel easier (\textit{``It make controlling easier, less effort"}). Many found that the system’s automatic adjustments to the path required less manual correction, making it easier to stay on course. Some participants also mentioned it being easier to use with the delay (\textit{``Changing the path is easier than controlling the robots specially due to the delay in the system"}).

While locus of control did not seem to influence personal preference, it is possible that other individual factors, such as comfort with technology, or prior experience with similar systems, could explain why users preferred one method over the other. 
Future research could explore what kinds of individual differences contribute to users' preferences, pitting increased control against prioritizing ease of use, or vice versa. Furthermore, \emph{adaptive control strategies that allow users to dynamically adjust the level of automation} is an interesting direction to explore, which provide flexibility based on individual preferences and task complexity.


Finally, our study confirms that introducing shared control does not degrade performance, and the findings suggest that it is a viable alternative to control switching, particularly for users who prioritize ease of use over direct control. However, additional refinements are needed to provide true benefits beyond what is offered by current methods. 
\bibliographystyle{IEEEtran}
\bibliography{bib}
\end{document}